\documentclass[11pt,a4paper]{article}
\usepackage{times,latexsym}
\usepackage{url}
\usepackage[T1]{fontenc}

\usepackage{times,latexsym}

\usepackage{placeins}
\usepackage{amsmath}
\usepackage[T1]{fontenc}
\usepackage{amssymb}
\usepackage{bbm}
\usepackage{graphicx}
\usepackage{booktabs}
\usepackage{mathtools}

\usepackage[acceptedWithA]{tacl2021v1}

\usepackage{xspace,mfirstuc,tabulary}

\newif\iftaclinstructions
\taclinstructionsfalse 
\iftaclinstructions
\renewcommand{\confidential}{}
\renewcommand{\anonsubtext}{(No author info supplied here, for consistency with
TACL-submission anonymization requirements)}
\newcommand{\instr}
\fi

\iftaclpubformat 

\else

\fi

\title{Post-Routing Arithmetic in Llama-3: Last-Token Result Writing and Rotation-Structured Digit Directions}

\author{
  Yao Yan \\
  Chongqing Normal University \\
  Chongqing Key Lab of Cognitive Intelligence and Intelligent Finance \\
  Chongqing, China \\
  \texttt{yaoyan@stu.cqnu.edu.cn}
}

\date{}

\begin{document}
\maketitle

\begingroup
\renewcommand\thefootnote{}\footnote{If this paper is accepted, we will release the code publicly.}
\addtocounter{footnote}{-1}
\endgroup

\begin{abstract}
  We study three-digit addition in Meta-Llama-3-8B (base) under a one-token readout to characterize how
  arithmetic answers are finalized after cross-token routing becomes causally irrelevant.
  Causal residual patching and cumulative attention ablations localize a sharp boundary near layer~17:
  beyond it, the decoded sum is controlled almost entirely by the last input token and late-layer self-attention
  is largely dispensable.
  In this post-routing regime, digit(-sum) direction dictionaries vary with a next-higher-digit context but are
  well-related by an approximately orthogonal map inside a shared low-rank subspace (low-rank Procrustes alignment).
  Causal digit editing matches this geometry: naive cross-context transfer fails, while rotating directions through the
  learned map restores strict counterfactual edits; negative controls do not recover.
  \end{abstract}

\section{Introduction}
\label{sec:intro}

Large language models (LLMs) solve multi-digit addition with near-perfect accuracy in-context, but we still lack a
mechanistic account of how a decoder-only transformer finalizes an arithmetic answer once attention-based routing
is no longer causally decisive.
We ask: after routing ends, where is the answer represented, and what writes it into the next-token prediction?

We study three-digit addition in Meta-Llama-3-8B (base) using a deliberately simple readout:
prompts of the form \texttt{"Calculate: \{a\}+\{b\} = "} followed by greedy decoding of a single next token,
parsed as an integer (Sec.~\ref{sec:setup}).
This one-token protocol makes causal localization sharp and allows direct intervention tests on
\texttt{resid\_post}.

We find a sharp \emph{result-writing boundary} near layer~16 (Sec.~\ref{sec:localization}):
cross-sample residual patching shifts causal control from non-last tokens to the last input token, and cumulative
attention ablation indicates that self-attention is largely dispensable beyond this boundary.
In the resulting post-routing regime, digit(-sum) direction dictionaries are strongly context-conditioned yet structured:
across shifts in a next-higher-digit nuisance context, within-context dictionaries are related by an approximately
orthogonal re-parameterization inside a shared low-rank core (Sec.~\ref{sec:ctx_rotations_digit_sum}).
We causally validate the corresponding prediction: cross-context transfer of directions collapses, while applying the
learned rotator restores strict counterfactual digit edits, with negative controls failing to recover
(Sec.~\ref{sec:digit_axis_editing}).

\paragraph{Contributions.}
\begin{itemize}
  \item We causally localize a sharp result-writing boundary near layer~16 where three-digit addition becomes
  effectively last-token-only and late-layer attention is largely dispensable.
  \item We characterize a context-conditioned but rotation-like structure in digit(-sum) direction dictionaries,
  captured by low-rank Procrustes alignment in a shared low-rank core.
  \item We provide causal validation via strict counterfactual digit editing: rotated directions recover cross-context
  editability while transfer fails, and negative controls do not recover.
\end{itemize}

\section{Related Work}
\label{sec:related}

\paragraph{Arithmetic and algorithmic reasoning.}\citep{anthropic2025tracing_thoughts,mamidanna-etal-2025-one,DBLP:conf/iclr/NikankinRMB25,quirke2024understanding,DBLP:conf/coling/ZhuDS25}
Mechanistic analyses of synthetic arithmetic often reveal structured internal algorithms and representations.\citep{DBLP:conf/emnlp/StolfoBS23,DBLP:conf/nips/ZhouFSJ24}
Work on modular arithmetic characterizes Fourier-like mechanisms and attention-mediated computation
\citep{nanda2023progress}. For multi-digit addition, recent studies identify digit-wise structure and carry-related
phenomena in transformer activations and circuits \citep{kruthoff2024carryingalgorithmtransformers,quirke2024understanding,DBLP:journals/corr/abs-2404-15255,DBLP:conf/iclr/ZhangN24}.
We focus specifically on late-stage answer finalization in a large decoder-only LLM under a one-token readout.

\paragraph{Causal interventions and activation editing.}
Beyond correlational probing \citep{DBLP:journals/coling/Belinkov22}, causal methods intervene on activations via
patching, mediation, and tracing \citep{DBLP:conf/emnlp/StolfoBS23,anthropic2025circuit_tracing}.
A complementary line studies low-dimensional functional directions as controllable edits (e.g., rank-one updates and
steering vectors) \citep{DBLP:conf/nips/MengBAB22,DBLP:conf/nips/ArditiOSPPGN24,chan2022causal}.
We combine these ideas to analyze a post-routing last-token regime and show that context-dependent direction families
are related by low-rank orthogonal alignment.

\section{Experimental Setup}
\label{sec:setup}

\paragraph{Task distribution.}
We study three-digit addition with operands sampled uniformly from $a,b\in\{1,\dots,999\}$.
Unless stated otherwise, we restrict to instances whose gold sum $s=a+b$ lies in $[200,999]$ so the result
is a three-digit integer and matches our one-token numeric readout.

\paragraph{Prompt templates.}
Our default prompt is \texttt{"Calculate: \{a\}+\{b\} = "} (note the trailing space).

\paragraph{Model and activations.}
We use Meta-Llama-3-8B (base) \citep{dubey2024llama}in BF16 without quantization.
Interventions operate on transformer-block outputs (\texttt{resid\_post}).
Throughout, ``layer $l$'' refers to the output of block $l\in\{1,\dots,32\}$.

\paragraph{One-token readout and parsing.}
We append a trailing space in the prompt so that the answer begins at a fresh token boundary.
With the Meta-Llama-3 tokenizer used in our experiments, every base-10 integer in $[0,1000]$ (in the
whitespace-prefixed form induced by our template) is represented as a single token; accordingly, we greedily
decode exactly the first next token and treat it as the model's numeric answer.

\paragraph{Baseline and evaluation convention.}
By default we report baseline (no-intervention) performance on the same evaluation distribution;
baseline strict accuracy is typically $\approx 99\%$ under this one-token protocol.
When an experiment conditions on baseline correctness (e.g., for scale selection), we state it explicitly.

\section{Causal Localization of the Result-Writing Boundary}
\label{sec:localization}

All later experiments assume a \emph{post-routing last-token regime} in which (i) the decoded arithmetic result is
causally controlled by the last input token's residual stream and (ii) late-layer self-attention is no longer necessary.\citep{mamidanna-etal-2025-one}
We establish this regime with two targeted causal tests at \texttt{resid\_post} under the one-token readout
(Sec.~\ref{sec:setup}): \textbf{(1) cross-sample residual patching}, which asks \emph{where} result semantics are stored,
and \textbf{(2) cumulative attention ablation}, which asks whether attention is still required once the boundary is reached.
Implementation details, per-layer curves, and auxiliary diagnostics are deferred to Appendix~\ref{app:localization_details}.

\subsection{Cross-sample residual patching localizes answer semantics}
\label{sec:method_cross_sample_patching}

We form baseline-correct source--target prompt pairs and patch \texttt{resid\_post} at a chosen layer $L$ by overwriting
either (i) the last-token row only (\emph{last-token patch}) or (ii) all non-last rows (\emph{non-last patch}), then decode
the single next token and parse it as an integer (Sec.~\ref{sec:setup})\citep{DBLP:conf/iclr/ZhangN24}. We measure \textbf{strict transfer}:
the patched target output equals the \emph{source} gold sum (parse failures count as failures).

Table~\ref{tab:cross-sample-phase} shows a sharp phase transition: early layers are controlled by non-last tokens, while
beyond layer~16 the answer becomes overwriteable almost entirely via the last-token residual state.

\begin{table}[t]
  \centering
  \small
  \begin{tabular}{lcc}
  \toprule
  Layer range
  & Last-token strict $\uparrow$
  & Non-last strict $\uparrow$ \\
  \midrule
  0--16   & 0.00 & $\approx$0.99 \\
  17--32  & 1.00 & $\approx$0.00 \\
  \bottomrule
  \end{tabular}
  \caption{
  Cross-sample \texttt{resid\_post} patching (strict transfer).
  Early layers are controlled by non-last tokens; after the transition near layer~16, overwriting only the last-token
  residual transfers the source answer, while overwriting non-last tokens has negligible effect.
  Full per-layer curves and additional diagnostics appear in Appendix~\ref{app:localization_details}.
  }
  \label{tab:cross-sample-phase}
\end{table}

\paragraph{Result-writing boundary.}
We define the \emph{result-writing boundary} as the depth at which causal control shifts from non-last tokens to the last
input token. For Llama-3-8B on three-digit addition under our one-token readout, this boundary is sharply localized to
\textbf{layer~16}.

\subsection{Attention ablation shows attention is dispensable beyond the boundary}
\label{sec:method_attention_ablation}

Residual patching localizes \emph{where} result semantics live, but does not by itself prove that self-attention is no
longer required after the boundary. We therefore perform \emph{cumulative attention ablation}: for a start layer $s$, we
ablate attention in the suffix $[s,32]$ by zeroing the attention-sublayer output in every ablated block, i.e.,
$\mathrm{attn\_out}^{(l)}(t)\leftarrow 0$ for all positions $t$ and all $l\in[s,32]$ (details in
Appendix~\ref{app:localization_details}), leaving the residual connection, MLP, and layernorm computations unchanged,
and evaluate strict accuracy on a held-out set filtered to baseline-correct examples under the same one-token protocol.

Table~\ref{tab:attn-ablation} corroborates the boundary: ablating attention starting at or before layer~16 collapses
accuracy, whereas ablating only layers \emph{after} the boundary (start $s\ge 17$) leaves accuracy essentially unchanged.
We provide the full accuracy-vs-start-layer curve with confidence intervals in Appendix~\ref{app:localization_details}.

\paragraph{Takeaway.}
Together, cross-sample patching and cumulative attention ablation establish a clean post-routing last-token regime:
\textbf{beyond layer~16}, three-digit addition is effectively last-token-only and late computation can be modeled as
transformations of the last-token residual stream. Accordingly, all subsequent causal edits operate on the last-token
\texttt{resid\_post} state at layers $l\ge 17$, unless otherwise stated.

\section{Methodology}
\label{sec:method_functional_vectors}

All experiments operate in the post-routing, last-token regime established in Sec.~\ref{sec:localization}.
Our intervention pipeline is: (i) learn simple \emph{functional vectors} from last-token states via
diff-of-means (DoM), (ii) when these directions vary with a nuisance context, align context-conditioned
\emph{dictionaries} across contexts using a low-rank orthogonal Procrustes map, and (iii) edit the last-token
\texttt{resid\_post} at a chosen layer and evaluate strict counterfactual success under a one-token readout
(Sec.~\ref{sec:setup}). Full implementation details are deferred to Appendix~\ref{app:method_details}.

\begin{table}[t]
  \centering
  \small
  \begin{tabular}{c c}
  \toprule
  Ablated Layers & Accuracy (\%) \\
  \midrule
  None (baseline) & 100.0 \\
  \midrule
  17--32 & $\approx$0.0 \\
  18--32 & $\ge$99.9 \\
  \bottomrule
  \end{tabular}
  \caption{
    Accuracy on baseline-correct subset (thus baseline=100\%).Strict accuracy after cumulative attention ablation.
  Attention is causally necessary through the result-writing boundary, but becomes dispensable in the post-routing regime
  (layers $\ge 17$) under the one-token readout.
  Full curves and confidence intervals are in Appendix~\ref{app:localization_details}.
  }
  \label{tab:attn-ablation}
  \end{table}

\subsection{Learning functional vectors (DoM)}
\label{sec:method_learn_functional}

Let $h^l_{\mathrm{last}}\in\mathbb{R}^d$ denote the last-token residual-stream state at layer $l$.
For a binary attribute $a$ (e.g., ``result hundreds digit equals $7$''), we define the unit DoM direction
\begin{equation}
\label{eq:dom_maintext}
\hat u^{(a)}_l
=
\frac{
\mathbb{E}\!\left[h^l_{\mathrm{last}} \mid a\right]
-
\mathbb{E}\!\left[h^l_{\mathrm{last}} \mid \neg a\right]
}{
\left\|
\mathbb{E}\!\left[h^l_{\mathrm{last}} \mid a\right]
-
\mathbb{E}\!\left[h^l_{\mathrm{last}} \mid \neg a\right]
\right\|_2
}.
\end{equation}
Expectations are estimated by sample averages over synthetic prompts (Sec.~\ref{sec:setup}).

\paragraph{Within-condition negatives.}
When learning directions for values under a fixed nuisance context $c$ (below), we construct negatives
\emph{within the same $c$} to prevent the direction from conflating the value with the context itself.

\subsection{Context-conditioned dictionaries and low-rank alignment}
\label{sec:method_rotated_alignment}

Many digit(-sum) directions are strongly \emph{context-conditioned}: the appropriate direction depends on a
nuisance context $c$ (e.g., next-higher digit context).
For discrete contexts $c\in\mathcal{C}$ and values $v\in\mathcal{V}$, we form within-context one-vs-rest
DoM directions:
\begin{equation}
\label{eq:ctx_dom_dict_method}
u^{(l)}_{c,v}
\;\coloneqq\;
\frac{
\mathbb{E}\!\left[h^l_{\mathrm{last}} \mid c,\; v\right]
-
\mathbb{E}\!\left[h^l_{\mathrm{last}} \mid c,\; v'\neq v\right]
}{
\left\|
\mathbb{E}\!\left[h^l_{\mathrm{last}} \mid c,\; v\right]
-
\mathbb{E}\!\left[h^l_{\mathrm{last}} \mid c,\; v'\neq v\right]
\right\|_2
}.
\end{equation}

\paragraph{Low-rank orthogonal Procrustes (summary).}
Stacking $\{u^{(l)}_{c,v}\}_{v\in\mathcal{V}}$ yields a value-indexed dictionary per context.
We compute a rank-$r$ basis for each context dictionary (e.g., by SVD) and estimate an orthogonal map
between the resulting coordinate systems by minimizing a Frobenius alignment loss:
\begin{equation}
\label{eq:procrustes_maintext}
R^{(c_{\mathrm{ref}}\rightarrow c)}_l
=
\arg\min_{R^\top R=I}
\left\|
\Phi^{(c_{\mathrm{ref}})}_l R - \Phi^{(c)}_l
\right\|_F .
\end{equation}
We then rotate reference-context directions into the target context and use them for editing.
We report alignment quality using (i) unaligned value-matched cosine, (ii) Procrustes-aligned cosine,
and (iii) a relative reconstruction error; precise definitions and additional controls appear in
Appendix~\ref{app:alignment_details}.

\subsection{Vector editing operators}
\label{sec:method_edit_ops}

All edits modify $h^l_{\mathrm{last}}$ at a chosen layer $l$ and then run the remaining layers unmodified,
followed by a greedy one-token readout (Sec.~\ref{sec:setup}).
Most experiments use a constant remove-and-add translation:
\begin{equation}
\label{eq:remove_add_unit}
h'^l_{\mathrm{last}}
=
h^l_{\mathrm{last}}
-\beta\,\hat u^{(a)}_l
+\alpha\,\hat u^{(b)}_l,
\qquad
\alpha,\beta\ge 0,
\end{equation}
where $a$ is the source attribute and $b$ is the target attribute.
We also use a norm-calibrated parameterization to make scales comparable across layers/attributes
(Appendix~\ref{app:edit_details}).

\subsection{Hyperparameter selection and reporting protocol}
\label{sec:method_selection_reporting}

We avoid per-pair tuning by selecting $(\alpha,\beta)$ on a small \emph{anchor} subset of edit pairs at each layer
(and, when relevant, each context/mode) using a fixed discrete grid, then freezing the chosen scales for the full sweep.
We evaluate all edit pairs with fixed scales on a held-out set and aggregate using weighted means with weights equal to
the number of valid evaluated examples $n_{\mathrm{total}}$.
For strict success and preservation rates we report Wilson 95\% confidence intervals; with multiple random seeds we pool
counts across seeds (equivalently weighting by $n_{\mathrm{total}}$).

\section{Context-Conditioned Rotations of Digit(-Sum) Direction Dictionaries}
\label{sec:ctx_rotations_digit_sum}

\paragraph{Motivation and phenomenon.}
Functional directions for a fixed digit(-sum) attribute often fail to transfer under shifts in a
\emph{nuisance context} variable (e.g., the next-higher digit context).
In the post-routing regime (Sec.~\ref{sec:localization}), we find that this failure is \emph{structured}:
for a fixed layer $L$, the within-context direction dictionary varies with context, yet the change is well
approximated by an \emph{approximately orthogonal} re-parameterization inside a shared low-rank core.
We quantify this using the dictionary alignment protocol and metrics introduced in
Appendix~\ref{app:alignment_details}, and refer to the resulting structure as
\emph{dictionary-level isomorphism across contexts}.

\subsection{Two digit-shifted settings}
\label{sec:ctx_rotations_settings}

We instantiate the context-conditioned dictionary construction in two adjacent digit places.
Across settings, $k$ denotes the \emph{focal digit(-sum) attribute} (either a digit-sum bucket or a digit-value bucket),
while $c$ denotes a \emph{next-higher-digit nuisance context} that shifts the representation but does not change the focal label.

\paragraph{(i) Ones-sum dictionaries across stripped-tens contexts.}
Define the focal ones-sum bucket
\begin{equation}
k ~\coloneqq~ (a \bmod 10) + (b \bmod 10) \in \{0,\dots,18\},
\end{equation}
and the stripped-tens context (pre-carry)
\begin{equation}
c ~=~ T ~\coloneqq~
\bigl(\lfloor a/10\rfloor + \lfloor b/10\rfloor\bigr) \bmod 10 \in \{0,\dots,9\}.
\end{equation}
Here $\mathcal{V}=\{0,\dots,18\}$.

\begin{figure*}[t]
  \centering
  \includegraphics[width=0.98\textwidth]{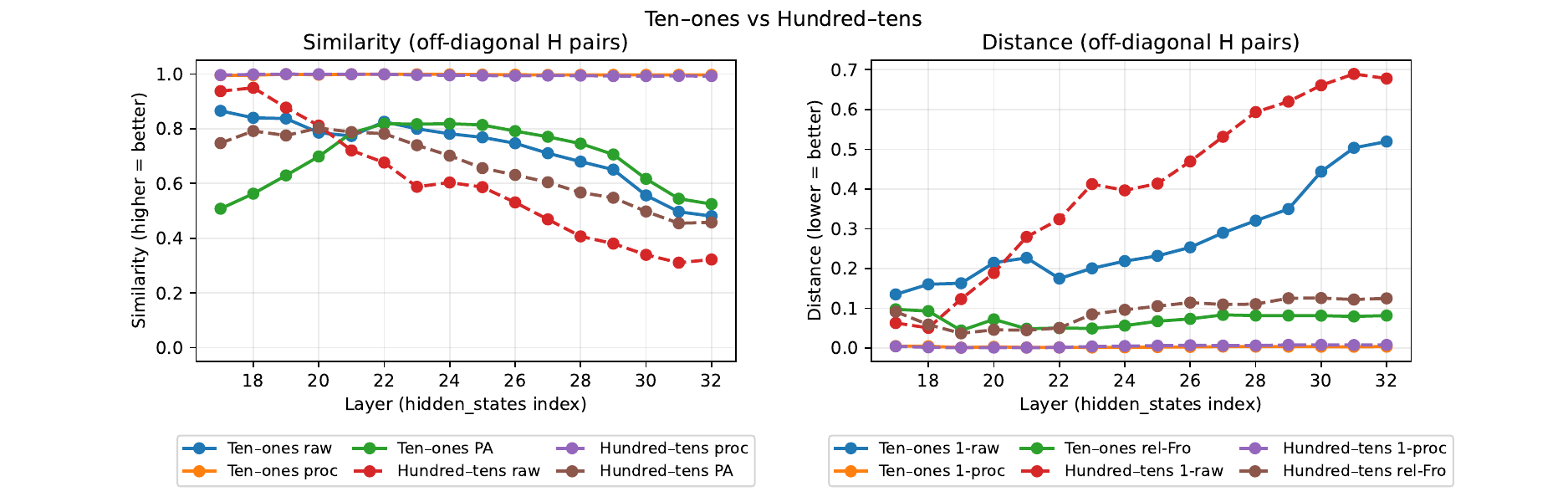}
  \vspace{-6pt}
  \caption{\textbf{Layerwise context-conditioned dictionary isomorphism (two digit-shifted settings).}
  For each layer $L$, metrics are averaged over off-diagonal context pairs $(c_1,c_2)$.
  \textbf{(a)} Unaligned cosine decreases with depth, indicating non-invariance of individual bucket directions.
  \textbf{(b)} After low-rank orthogonal Procrustes alignment (Sec.~\ref{app:alignment_details}),
  aligned cosine remains near unity while $\mathrm{RelFro}$ stays small, consistent with a rotation-like
  re-parameterization of a shared low-rank core.
  Solid: stripped-tens context $\rightarrow$ ones-sum; dashed: result-hundreds-digit context $\rightarrow$ result-tens digit.}
  \label{fig:dict_iso_layerwise}
\end{figure*}

\begin{figure}[t]
  \centering
  \includegraphics[width=\columnwidth]{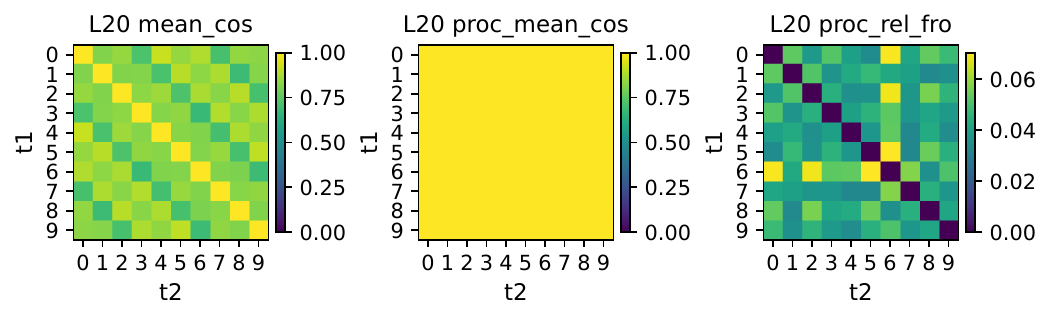}\\[-2pt]
  \caption{\textbf{Single-layer anchor: context-conditioned rotation at a fixed depth.}
  At a representative post-routing layer, unaligned similarity between context dictionaries varies
  substantially (top), while low-rank Procrustes alignment collapses similarity to near unity (bottom).}
  \label{fig:dict_anchor_singlecol}
\end{figure}

\subsection{Results: dictionary-level isomorphism across contexts}
\label{sec:ctx_rotations_results}

\paragraph{Layerwise summary.}
Figure~\ref{fig:dict_iso_layerwise} shows a layerwise sweep of cross-context dictionary similarity
for both digit-shifted settings, averaging over \emph{off-diagonal} context pairs $(c_1,c_2)$ with $c_1\neq c_2$.
Unaligned similarity $\mathrm{Cos}_{\mathrm{unaligned}}$ decreases with depth, indicating that individual
bucket directions are not invariant under context shifts.
However, after low-rank Procrustes alignment (details in Appendix~\ref{app:alignment_details}),
$\mathrm{Cos}_{\mathrm{proc}}$ remains near unity while $\mathrm{RelFro}$ stays small,
consistent with an approximately orthogonal re-parameterization inside a shared low-rank core.

\paragraph{A concrete visual anchor.}
Figure~\ref{fig:dict_anchor_singlecol} shows a representative post-routing layer in the ones-sum setting:
unaligned similarity varies substantially across context pairs, whereas Procrustes alignment collapses
similarity to near unity, illustrating the same effect at a fixed depth.

\paragraph{Takeaway and causal prediction.}
This section establishes a structured geometric claim: digit(-sum) direction dictionaries vary with nuisance context,
yet are related by an approximately orthogonal change of coordinates in a shared low-rank core.
In the next section (Sec.~\ref{sec:digit_axis_causal_rotation}), we test the corresponding \emph{causal prediction}:
directions learned under one context should fail when applied under a shifted context (\textsc{transfer}),
but should recover after applying the learned rotator (\textsc{rotated}), while negative controls should not recover.

\section{One-Dimensional Result-Digit Editing Axes}
\label{sec:digit_axis_editing}

We now give a causal test of the context-conditioned rotation claim
(Sec.~\ref{sec:ctx_rotations_digit_sum}) in the post-routing last-token regime
(Sec.~\ref{sec:localization}).

The operational prediction is simple:
directions learned under one nuisance context typically fail to edit under a shifted context
(\textsc{transfer}), but should recover after applying the learned low-rank rotator
(\textsc{rotated}), approaching the within-context upper bound (\textsc{direct});
negative controls should not recover.
All interventions modify last-token \texttt{resid\_post} at layer $l$ and are evaluated by strict
numeric counterfactuals under the one-token readout (Sec.~\ref{sec:setup}).

\begin{figure*}[t]
  \centering
  \includegraphics[width=0.98\textwidth]{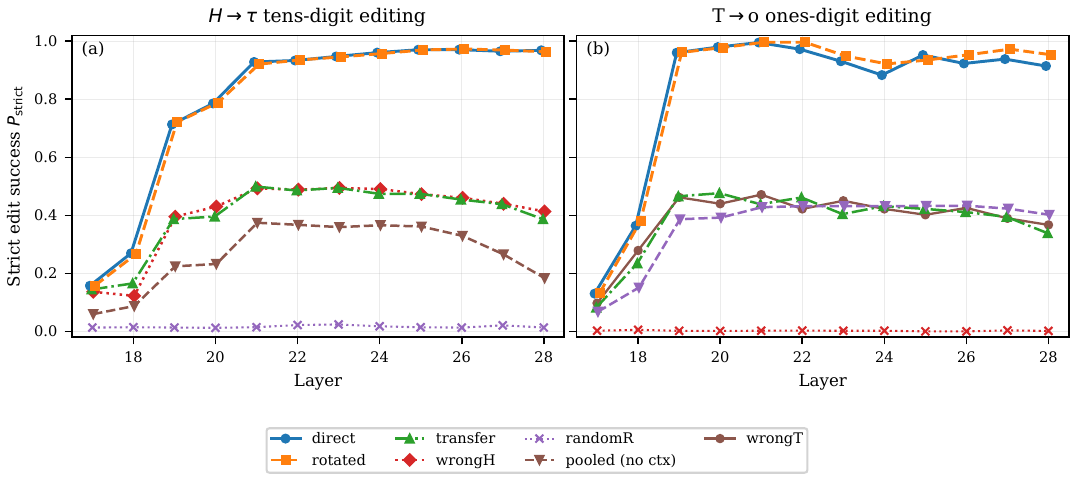}
  \vspace{-6pt}
  \caption{\textbf{Rotation is necessary for cross-context digit editing.}
  Strict counterfactual success vs.\ layer (baseline-correct subset).
  \textsc{rotated} recovers toward \textsc{direct}, while \textsc{transfer} collapses;
  \textsc{wrong-condition} and \textsc{random-$R$} fail to recover.}
  \label{fig:rotation_causal_main}
\end{figure*}
\subsection{Setup: strict counterfactual targets, diagnostics, and modes}
\label{sec:digit_axis_setup}

\paragraph{Digit edit objective.}
Let $s=a+b=100h+10x+y$ be the gold three-digit sum.
A digit edit selects a place $p\in\{\textsc{hundreds},\textsc{tens},\textsc{ones}\}$ and a value change
$(v\!\to\!v')$ at that place.
We allow carry/borrow propagation and define the strict target by the induced numeric offset:
\begin{equation}
  \label{eq:strict_target}
  \begin{aligned}
  \Delta(p; v\!\to\!v') &= (v'-v)\cdot 10^{k(p)}, \\
  k(\textsc{hundreds}) &= 2,\quad k(\textsc{tens})=1,\quad k(\textsc{ones})=0, \\
  s_{\mathrm{cf}} &= s + \Delta(p; v\!\to\!v').
  \end{aligned}
  \end{equation}
  We report \textbf{strict} success iff the decoded integer equals $s_{\mathrm{cf}}$
  (parse failures count as failures).
  
\paragraph{Illustrative example (tens-digit edit).}
Consider an instance whose baseline decoded (gold) sum is the three-digit integer
$s=423$, so $(h,x,y)=(4,2,3)$.
We choose place $p=\textsc{tens}$ and value change $(v\!\to\!v')=(2\!\to\!5)$.
By Eq.~\eqref{eq:strict_target}, the induced offset is
$\Delta = (5-2)\cdot 10 = 30$, hence the strict counterfactual target is
$s_{\mathrm{cf}} = 423 + 30 = 453$ (no carry/borrow occurs in this example).

At a post-routing layer $l$, let $h^l_{\mathrm{last}}$ denote the last-token
\texttt{resid\_post} state for this prompt.
Let $\hat u^{(H=4,\;t=2)}_l$ and $\hat u^{(H=4,\;t=5)}_l$ be the unit DoM directions
for the \textsc{tens} digit value $t\in\{2,5\}$ under the hundreds-digit context $H=4$
(Eq.~\eqref{eq:ctx_dom_dict_method}).
We apply the remove-and-add edit (Eq.~\eqref{eq:remove_add_unit}):
\begin{align}
h'^l_{\mathrm{last}}
&=
h^l_{\mathrm{last}}
-\beta\,\hat u^{(H=4,\;t=2)}_l
+\alpha\,\hat u^{(H=4,\;t=5)}_l,
\label{eq:illustrative_tens_edit}\\
&\qquad \alpha,\beta\ge 0,
\nonumber
\end{align}
then run the remaining blocks unmodified and decode with the one-token readout.
We expect this intervention to causally shift the model's decoded output from $423$ to $453$,
i.e., replacing the tens digit ``$2$'' with ``$5$'' while preserving the hundreds and ones digits.

  \begin{figure*}[t]
  \centering
  \includegraphics[width=\textwidth]{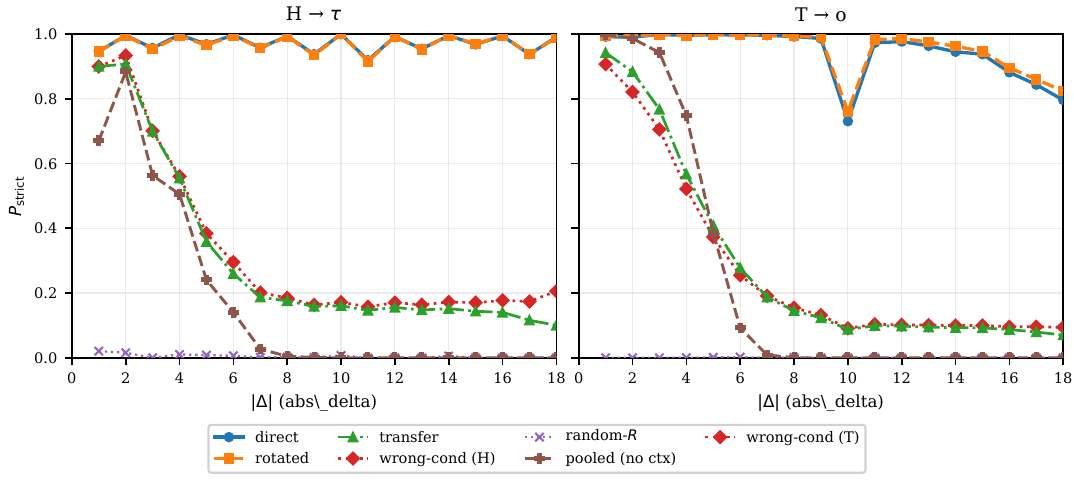}
  \vspace{-4pt}
  \caption{\textbf{Edit-magnitude sensitivity across two settings.}
  Strict success vs.\ $|\Delta|$ (layers 24--26, weighted by $n_{\mathrm{total}}$).
  Left: $H\!\rightarrow\!t$. Right: $T\!\rightarrow\!o$.
  \textsc{transfer} degrades with $|\Delta|$, while \textsc{rotated} remains close to \textsc{direct};
  \textsc{random-$R$} and wrong-condition controls do not recover.}
  \label{fig:delta_sensitivity_Ht}
\end{figure*}
  
\paragraph{Diagnostics.}
In addition to strict success, we report (i) \textbf{parse rate} and (ii) non-target digit preservation:
$\mathrm{Preserve}_{\neg p}=1$ iff the two non-target digits match $(h,x,y)$.

\paragraph{Modes and controls.}
We compare:
\textsc{direct} (learn+edit within the eval context),
\textsc{transfer} (reuse a source-context direction without adjustment),
\textsc{rotated} (rotate the source direction into the eval context; Sec.~\ref{app:alignment_details}).
Controls are \textsc{wrong-condition} (mismatched rotator) and \textsc{random-$R$}
(random orthogonal map at the same rank), ruling out ``any rotation works'' explanations.

\begin{figure}[t]
  \centering
  \includegraphics[width=\linewidth]{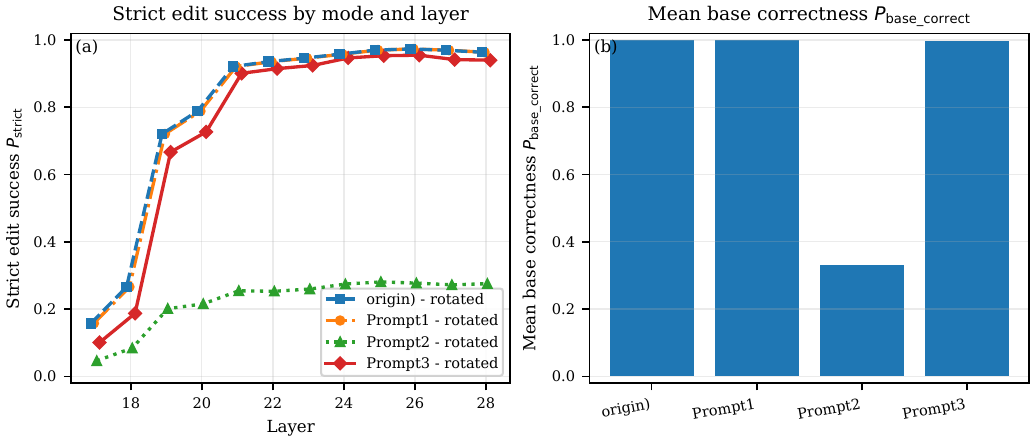}
  \caption{
  Cross-prompt transport of digit-editing interventions.
  \textbf{(a)} Strict edit success vs.\ layer under the canonical template and three paraphrases.
  \textbf{(b)} Right: bar plot of answer accuracy under each prompt.
  }
  \label{fig:digit_axis_prompt_transfer}
\end{figure}

\subsection{Main causal test: rotation enables cross-context digit editing}
\label{sec:digit_axis_causal_rotation}

We evaluate two context-shifted settings aligned with Sec.~\ref{sec:ctx_rotations_settings}:
(i) ones-digit editing under stripped-tens context shifts ($T$), and
(ii) tens-digit editing under hundreds-context shifts ($H$).
At layer $l$, we learn within-context value directions (Eq.~\ref{eq:ctx_dom_dict_method}),
estimate rank-$r$ Procrustes rotators between contexts (Sec.~\ref{app:alignment_details}),
and apply the same remove-and-add edit operator (Eq.~\ref{eq:remove_add_unit}) using either
direct, transferred, or rotated directions.
Figure~\ref{fig:rotation_causal_main} shows the predicted pattern:
\textsc{transfer} collapses under context shift, while \textsc{rotated} largely restores strict success
toward \textsc{direct}; controls do not recover.

\paragraph{Edit-magnitude sensitivity.}
To stress-test cross-context failure, we stratify by $|\Delta|=|v'-v|$ (larger edits induce more carry interactions).
Figure~\ref{fig:delta_sensitivity_Ht} aggregates strict success over representative post-routing layers (24--26):
\textsc{transfer} degrades sharply with $|\Delta|$, while \textsc{rotated} tracks \textsc{direct};
\textsc{random-$R$} remains near chance.

\subsection{Cross-prompt transport of digit-editing interventions}
\label{sec:digit_axis_prompt_transfer}

To test whether the learned directions encode result-digit semantics rather than template-specific artifacts,
we learn all directions (and rotators, when applicable) from a single canonical template:
\texttt{"Calculate: \{a\}+\{b\} = "}.
We then apply the same last-token interventions under three alternative paraphrases.

As shown in Fig.~\ref{fig:digit_axis_prompt_transfer}, Prompt1 and Prompt3 yield curves close to the origin,
whereas Prompt2 is markedly worse; we view this as an interesting extension and do not further analyze
prompt-sensitivity in this section.

\subsection{Takeaway}
\label{sec:digit_axis_by_place}

In the post-routing last-token regime, individual result digits admit near one-dimensional causal control:
(i) within-context digit edits succeed under strict numeric counterfactual evaluation,
(ii) cross-context reuse fails in a structured way, and (iii) the learned low-rank rotation restores editability,
while negative controls do not.
Together with cross-prompt transport, these results support interpreting the learned directions as
task-semantic result-digit representations in the last-token residual stream.

\section{Limitations}
\label{sec:limitations}

\begin{itemize}
  \item \textbf{Task and readout specificity.}
  Our findings are established for three-digit base-10 addition in a single model (Meta-Llama-3-8B, base) under a
  deliberately constrained \emph{one-token} numeric readout.
  The location and sharpness of the result-writing boundary, as well as the post-routing ``last-token-only'' behavior,
  may change for other arithmetic tasks (e.g., subtraction/multiplication), longer numbers, different tokenizers, or
  multi-token generation where later tokens can re-introduce routing dependencies.

  \item \textbf{Partial mechanistic coverage.}
  We focus on late-stage answer finalization \emph{after} cross-token routing becomes causally irrelevant.
  Our methods do not provide a full end-to-end explanation of how earlier layers compute carries and intermediate
  representations, nor do they fully characterize the circuits that route information into the last token before the
  boundary.

  \item \textbf{Intervention class limitations.}
  Our causal edits rely on simple diff-of-means functional vectors and low-rank orthogonal alignment.
  These tools may miss nonlinear structure, multi-direction interactions, or cases where the relevant subspace is not
  well-approximated by a shared low-rank core. Similarly, ``rotation-like'' alignment is an approximation and may break
  under harder distribution shifts.

  \item \textbf{Broader replication.}
  We do not yet know how these phenomena vary across model families, scales, training recipes, or fine-tuned/chat
  variants. Cross-model replication and systematic comparisons are required before treating the boundary and the
  low-rank alignment structure as general properties of decoder-only LLMs.
\end{itemize}

\section{Conclusion}
\label{sec:conclusion}

We presented a causal and geometric account of how Meta-Llama-3-8B (base) finalizes three-digit
addition once cross-token routing ceases to matter.
Using a one-token readout, cross-sample residual patching localizes a sharp result-writing boundary
near layer~17, where causal control of the decoded sum shifts from non-last tokens to the last input
token. Complementary cumulative attention ablations indicate that, beyond this boundary, late-layer
self-attention is largely dispensable, supporting a post-routing picture in which computation is
dominated by MLP-mediated updates to the last-token residual stream.

Within this post-routing regime, we find that digit(-sum) functional directions are strongly
context-conditioned but structured: dictionaries learned under different next-higher-digit contexts
are well-related by an approximately orthogonal re-parameterization inside a shared low-rank core,
captured by low-rank Procrustes alignment. We then causally validate the corresponding prediction in
digit editing: direct within-context edits succeed, naive cross-context transfer collapses, and
rotating directions through the learned alignment largely restores strict counterfactual performance,
while negative controls fail to recover. Together, these results support a mechanistic view in which
the model writes arithmetic results into the last token after routing ends, and the relevant
representations transform predictably across nuisance-context shifts.

A natural next step is to extend this framework beyond three-digit addition and beyond the specific
one-token readout, and to map the low-rank aligned subspaces to more explicit circuit components.

\bibliography{tacl2021}
\bibliographystyle{acl_natbib}

\appendix

\label{sec:appendix}

\section{Details for causal localization (patching and attention ablation)}
\label{app:localization_details}

We provide the exact interventions and the full curves referenced in Sec.~\ref{sec:localization}.

\paragraph{Setup and filtering.}
We use the prompt \texttt{"Calculate: \{a\}+\{b\} = "} with a trailing space and the shared one-token parsing protocol
(Sec.~\ref{sec:setup}). For cross-sample patching, we form source--target pairs with distinct gold sums and filter to
pairs where the unmodified model is baseline-correct on \emph{both} prompts (to avoid confounding from baseline errors).

\paragraph{Cross-sample patching (exact definition).}
Let $X=(x_1,\ldots,x_S)$ be the tokenized prompt and $\mathbf{r}^{(L)}(X)\in\mathbb{R}^{S\times d}$ be \texttt{resid\_post}
at layer $L$. Given $(X_{\mathrm{src}},X_{\mathrm{tgt}})$, we cache $\mathbf{r}^{(L)}(X_{\mathrm{src}})$ and re-run
$X_{\mathrm{tgt}}$ while overwriting either:

\begin{alignat}{2}
  \textbf{Last-token patch:}\quad
  & \mathbf{r}^{(L)}_{S}(X_{\mathrm{tgt}})
  && \leftarrow \mathbf{r}^{(L)}_{S}(X_{\mathrm{src}}), \\
  \textbf{Non-last patch:}\quad
  & \mathbf{r}^{(L)}_{i}(X_{\mathrm{tgt}})
  && \leftarrow \mathbf{r}^{(L)}_{i}(X_{\mathrm{src}}), \\
  &&& \text{for } i=1,\ldots,S-1. \notag
  \end{alignat}

We then run layers $L{+}1,\ldots,31$ normally and decode one token. \textbf{Strict transfer} counts success iff the
parsed output equals $s_{\mathrm{src}}$ (parse failures are failures).

\begin{figure}[t]
\centering
\includegraphics[width=\columnwidth]{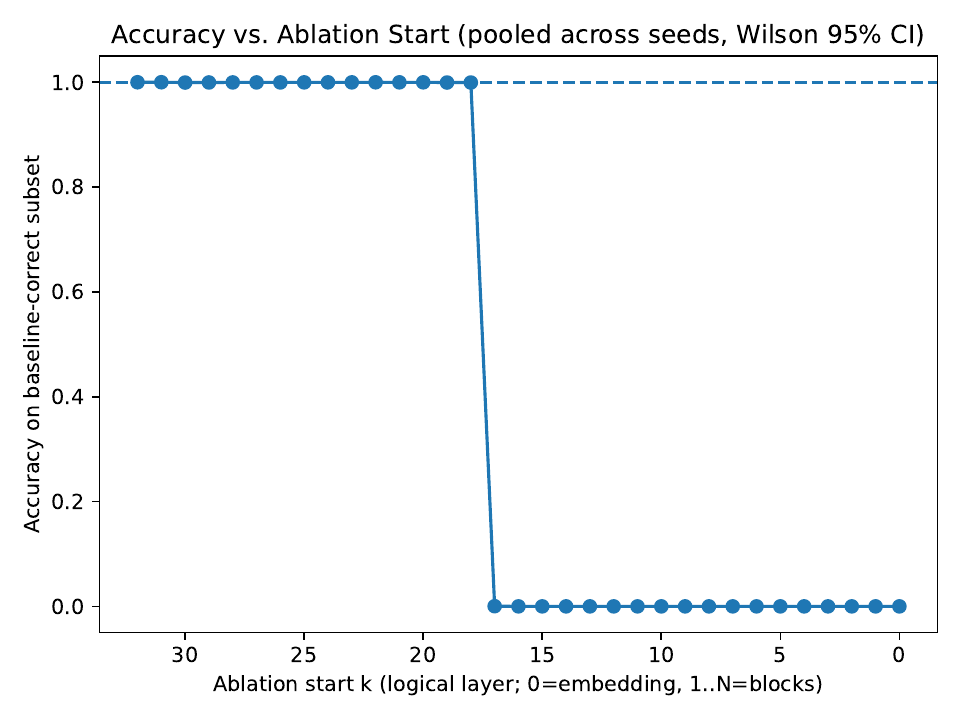}\\[-2pt]
\caption{
\textbf{Cross-sample patching curves.}
Strict transfer vs.\ layer for last-token and non-last patching on baseline-correct source--target pairs.
A sharp crossover occurs near layer~16.
}
\label{fig:patching_curve_full}
\end{figure}

\paragraph{Cumulative attention ablation and uncertainty.}
For a start layer $s$, we ablate attention outputs in all blocks $l\in[s,31]$ (MLP and residual path intact), then decode
with the same one-token protocol. We evaluate strict accuracy on a held-out set filtered to baseline-correct examples.
When using multiple seeds, we pool counts and report Wilson 95\% confidence intervals.

\section{Additional Results for Digit-Sum Dictionary Rotations}
\label{app:digit_sum_tables}

\subsection{Full layerwise summary tables}
\label{app:digit_sum_tables_layerwise}

\begin{table*}[t]
  \centering
  \small
  \setlength{\tabcolsep}{4pt}
  \renewcommand{\arraystretch}{1.0}

  \begin{tabular}{rccc@{\hspace{10pt}}ccc}
  \toprule
  & \multicolumn{3}{c}{ones-sum dict (stripped-tens)}
  & \multicolumn{3}{c}{tens-sum dict (stripped-hundreds)} \\
  \cmidrule(lr){2-4}\cmidrule(lr){5-7}
  Layer & Unaligned & Procrustes & RelFro & Unaligned & Procrustes & RelFro \\
  \midrule
  17 & $0.94 \pm 0.02$ & $1.00 \pm 0.00$ & 0.09 & $0.87 \pm 0.10$ & $1.00 \pm 0.00$ & 0.10 \\
  18 & $0.95 \pm 0.02$ & $1.00 \pm 0.00$ & 0.06 & $0.84 \pm 0.19$ & $1.00 \pm 0.00$ & 0.09 \\
  19 & $0.88 \pm 0.08$ & $1.00 \pm 0.00$ & 0.04 & $0.84 \pm 0.11$ & $1.00 \pm 0.00$ & 0.04 \\
  20 & $0.81 \pm 0.14$ & $1.00 \pm 0.00$ & 0.04 & $0.79 \pm 0.26$ & $1.00 \pm 0.00$ & 0.07 \\
  21 & $0.72 \pm 0.21$ & $1.00 \pm 0.00$ & 0.05 & $0.77 \pm 0.15$ & $1.00 \pm 0.00$ & 0.05 \\
  22 & $0.68 \pm 0.21$ & $1.00 \pm 0.00$ & 0.05 & $0.83 \pm 0.11$ & $1.00 \pm 0.00$ & 0.05 \\
  23 & $0.59 \pm 0.27$ & $1.00 \pm 0.01$ & 0.08 & $0.80 \pm 0.11$ & $1.00 \pm 0.00$ & 0.05 \\
  24 & $0.60 \pm 0.20$ & $0.99 \pm 0.01$ & 0.10 & $0.78 \pm 0.08$ & $1.00 \pm 0.00$ & 0.06 \\
  25 & $0.59 \pm 0.16$ & $0.99 \pm 0.00$ & 0.11 & $0.77 \pm 0.07$ & $1.00 \pm 0.00$ & 0.07 \\
  26 & $0.53 \pm 0.20$ & $0.99 \pm 0.01$ & 0.11 & $0.75 \pm 0.06$ & $1.00 \pm 0.00$ & 0.07 \\
  27 & $0.47 \pm 0.17$ & $0.99 \pm 0.00$ & 0.11 & $0.71 \pm 0.07$ & $1.00 \pm 0.00$ & 0.08 \\
  28 & $0.41 \pm 0.18$ & $0.99 \pm 0.00$ & 0.11 & $0.68 \pm 0.07$ & $1.00 \pm 0.00$ & 0.08 \\
  \bottomrule
  \end{tabular}

  \caption{\textbf{Full layerwise summary for digit-sum dictionary rotations.}
  Mean$\pm$std over context pairs for unaligned ($k$-aligned) cosine and Procrustes-aligned cosine,
  along with relative Frobenius reconstruction error (RelFro), reported for layers $L\in[17,28]$
  in the ones-sum (stripped-tens context) and tens-sum (stripped-hundreds context) settings.}
  \label{tab:dict_iso_to_ht_full}
\end{table*}

\section{Additional Method Details}
\label{app:method_details}

This appendix provides mathematical and implementation details for the methodology in
Sec.~\ref{sec:method_functional_vectors}, including conditional dictionary construction,
low-rank orthogonal alignment, rotated transfer, control modes, and norm-calibrated editing.

\subsection{DoM estimation details}
\label{app:dom_details}

We estimate expectations in Eqs.~\eqref{eq:dom_maintext} and \eqref{eq:ctx_dom_dict_method} by sample averages over
synthetic prompts (Sec.~\ref{sec:setup}). For conditional directions we always use within-condition negatives
(i.e., negatives share the same context $c$) to avoid conflating value with context.

\subsection{Conditional dictionaries and low-rank orthogonal alignment}
\label{app:alignment_details}

\paragraph{Conditional attributes.}
Let $c\in\mathcal{C}$ denote a condition/domain variable and $v\in\mathcal{V}$ denote the
edited attribute value.
At layer $l$, define the conditional DoM direction (one-vs-rest within fixed $c$):
\begin{equation}
\label{eq:conditional_dom_app}
\begin{aligned}
u^{(c,v)}_l
&\coloneqq
\mathbb{E}\!\left[h^{l}_{\mathrm{last}} \mid c,\; v\right]
-
\mathbb{E}\!\left[h^{l}_{\mathrm{last}} \mid c,\; v' \in \mathcal{V}\!\setminus\!\{v\}\right],\\
\hat u^{(c,v)}_l
&\coloneqq
\frac{u^{(c,v)}_l}{\|u^{(c,v)}_l\|_2}.
\end{aligned}
\end{equation}

\paragraph{Row matrices for alignment.}
For each condition $c$, stack row-normalized directions across values $v$:
\begin{equation}
\label{eq:U_matrix_app}
U^{(c)}_l
\coloneqq
\bigl[\hat u^{(c,v)}_l\bigr]_{v\in\mathcal{V}}
\in \mathbb{R}^{|\mathcal{V}|\times d}.
\end{equation}

\paragraph{Low-rank bases.}
To avoid estimating a full $d\times d$ rotation, we compute a rank-$r$ orthonormal basis
$B^{(c)}_l\in\mathbb{R}^{d\times r}$ for each condition (e.g., by taking the top-$r$
right singular vectors of $U^{(c)}_l$).

We denote the coordinates of a direction in this basis by
\begin{equation}
\label{eq:phi_app}
\begin{aligned}
\phi^{(c,v)}_l
&\coloneqq
\hat u^{(c,v)}_l B^{(c)}_l
\in \mathbb{R}^{r},\\
\Phi^{(c)}_l
&\coloneqq
U^{(c)}_l B^{(c)}_l
\in \mathbb{R}^{|\mathcal{V}|\times r}.
\end{aligned}
\end{equation}

\paragraph{Orthogonal Procrustes rotators.}
Given a reference condition $c_{\mathrm{ref}}$ and an evaluation condition $c$,
we estimate an orthogonal matrix $R^{(c_{\mathrm{ref}}\rightarrow c)}_l\in\mathbb{R}^{r\times r}$ by
\begin{equation}
\label{eq:procrustes_app}
R^{(c_{\mathrm{ref}}\rightarrow c)}_l
\coloneqq
\arg\min_{R^\top R = I}
\left\|
\Phi^{(c_{\mathrm{ref}})}_l R - \Phi^{(c)}_l
\right\|_F .
\end{equation}
In practice this is solved via the SVD of
$\bigl(\Phi^{(c_{\mathrm{ref}})}_l\bigr)^{\!\top}\Phi^{(c)}_l$.

\paragraph{Rotated transfer of functional directions.}
To transfer an attribute direction learned under $c_{\mathrm{ref}}$ to condition $c$,
we map it through aligned low-rank subspaces:
\begin{equation}
\label{eq:rotate_u_lowrank_app}
\begin{aligned}
z^{(c_{\mathrm{ref}}, v)}_l
&\coloneqq
\bigl(B^{(c_{\mathrm{ref}})}_l\bigr)^{\!\top}\hat u^{(c_{\mathrm{ref}}, v)}_l
\in \mathbb{R}^{r},\\
\bar u^{(c \leftarrow c_{\mathrm{ref}}, v)}_l
&\coloneqq
B^{(c)}_l\, R^{(c_{\mathrm{ref}}\rightarrow c)}_l\, z^{(c_{\mathrm{ref}}, v)}_l
\in \mathbb{R}^{d},\\
\widetilde u^{(c \leftarrow c_{\mathrm{ref}}, v)}_l
&\coloneqq
\frac{\bar u^{(c \leftarrow c_{\mathrm{ref}}, v)}_l}
{\left\|\bar u^{(c \leftarrow c_{\mathrm{ref}}, v)}_l\right\|_2}
\in \mathbb{R}^{d}.
\end{aligned}
\end{equation}
We use $\widetilde u$ in the same editing operators as direct directions.

\paragraph{Control modes.}
We report multiple modes that isolate direction quality versus alignment quality:
\begin{itemize}
  \item \textbf{direct:} use $\hat u^{(c,v)}_l$ learned under the evaluation condition $c$.
  \item \textbf{transfer:} use $\hat u^{(c_{\mathrm{ref}},v)}_l$ directly without rotation.
  \item \textbf{rotated:} use $\widetilde u^{(c \leftarrow c_{\mathrm{ref}}, v)}_l$ from Eq.~\eqref{eq:rotate_u_lowrank_app}.
  \item \textbf{wrong-condition:} rotate using $R^{(c_{\mathrm{ref}}\rightarrow c_{\mathrm{wrong}})}_l$ for some $c_{\mathrm{wrong}}\neq c$.
  \item \textbf{random-$R$:} rotate using a random orthogonal matrix in $\mathbb{R}^{r\times r}$.
\end{itemize}

\subsection{Editing operators and scale calibration}
\label{app:edit_details}

\paragraph{Additive steering.}
Given a unit direction $\hat u^{(a)}_l$ and scalar $\alpha\in\mathbb{R}$,
\begin{equation}
\label{eq:additive_app}
h'^{\,l}_{\mathrm{last}}
=
h^l_{\mathrm{last}} + \alpha \,\hat u^{(a)}_l .
\end{equation}

\paragraph{Constant remove-and-add editing.}
Given attributes $a$ (source) and $b$ (target),
\begin{equation}
\label{eq:remove_add_app}
h'^l_{\mathrm{last}}
=
h^l_{\mathrm{last}}
-\beta\,\hat u^{(a)}_l
+\alpha\,\hat u^{(b)}_l,
\qquad
\alpha,\beta\ge 0.
\end{equation}

\paragraph{Norm-calibrated parameterization.}
To improve comparability of scale parameters across layers/attributes, we also use
\begin{equation}
\label{eq:remove_add_unit_app}
h'^l_{\mathrm{last}}
=
h^l_{\mathrm{last}}
-\beta_s\,\lVert r^{(a)}_l\rVert_2\,\hat u^{(a)}_l
+\alpha_s\,\lVert r^{(b)}_l\rVert_2\,\hat u^{(b)}_l,
\end{equation}
where $\alpha_s,\beta_s$ are dimensionless scales and $r^{(a)}_l, r^{(b)}_l$ denote the
corresponding (unnormalized) DoM vectors before unit normalization.

\paragraph{Hyperparameter selection (implementation).}
Unless otherwise stated, we sweep a small grid of norm-calibrated scales
$\alpha_s,\beta_s \in \{0.5,1.0\}$, select the best-performing pair on a small
anchor subset, and then evaluate with fixed scales on a larger set.
We always report baseline (no-intervention) performance under the same parsing and strict metrics.

\subsection{Exact cross-context dictionary metrics}
\label{app:ctx_dict_metrics_exact}

Fix layer $l$. For each context $c$, stack unit directions over values to form a dictionary
\begin{equation}
\label{eq:app_U_dict}
U^{(l)}_{c}
\;\coloneqq\;
\begin{bmatrix}
u^{(l)}_{c,v_1}\\
\vdots\\
u^{(l)}_{c,v_{|\mathcal{V}|}}
\end{bmatrix}
\in\mathbb{R}^{|\mathcal{V}|\times d}.
\end{equation}

\paragraph{Unaligned cosine.}
For a context pair $(c_1,c_2)$,
\begin{equation}
\label{eq:app_cos_unaligned}
\mathrm{Cos}_{\mathrm{unaligned}}(c_1,c_2)
=\frac{1}{|\mathcal{V}|}\sum_{v\in\mathcal{V}}
\left\langle u^{(l)}_{c_1,v},\,u^{(l)}_{c_2,v}\right\rangle.
\end{equation}

\paragraph{Low-rank Procrustes alignment.}
Let $B^{(l)}_{c}\in\mathbb{R}^{d\times r}$ be the top-$r$ right singular vectors of $U^{(l)}_{c}$ and define
$X^{(l)}_{c}=U^{(l)}_{c}B^{(l)}_{c}\in\mathbb{R}^{|\mathcal{V}|\times r}$.
For $(c_1,c_2)$, solve
\begin{equation}
\label{eq:app_procrustes_coords}
R^{(l)}_{c_1\to c_2}
=
\arg\min_{R^\top R=I}\left\|X^{(l)}_{c_1}R-X^{(l)}_{c_2}\right\|_F,
\end{equation}
and reconstruct an aligned dictionary in $c_2$'s ambient coordinates:
\begin{equation}
\label{eq:app_aligned_recon}
U^{(l)}_{c_1\to c_2}
=
\left(U^{(l)}_{c_1}B^{(l)}_{c_1}\right)
R^{(l)}_{c_1\to c_2}
\left(B^{(l)}_{c_2}\right)^\top.
\end{equation}
Row-normalize $U^{(l)}_{c_1\to c_2}$ before cosine computation.

\paragraph{Aligned cosine and reconstruction error.}
Let $\tilde u^{(l)}_{c_1\to c_2,v}$ be the row-normalized $v$-th row of $U^{(l)}_{c_1\to c_2}$. Then
\begin{align}
\label{eq:app_cos_proc}
\mathrm{Cos}_{\mathrm{proc}}(c_1,c_2)
&=\frac{1}{|\mathcal{V}|}\sum_{v\in\mathcal{V}}
\left\langle \tilde u^{(l)}_{c_1\to c_2,v},\,u^{(l)}_{c_2,v}\right\rangle,\\
\label{eq:app_relfro}
\mathrm{RelFro}(c_1,c_2)
&=\frac{\left\|U^{(l)}_{c_1\to c_2}-U^{(l)}_{c_2}\right\|_F}{\left\|U^{(l)}_{c_2}\right\|_F}.
\end{align}

\paragraph{Averaging over context pairs.}
For layerwise summaries, we average each metric over off-diagonal context pairs $(c_1,c_2)$ with $c_1\neq c_2$.

\end{document}